\documentclass{egpubl}
\CGFccby
\usepackage[T1]{fontenc}
\usepackage{dfadobe}   
\usepackage{cite}  % comment out for biblatex with backend=biber 
\usepackage{epigraph} 
\usepackage{amsmath} 
\usepackage{xspace}
\usepackage{pifont}
\usepackage{booktabs}
\usepackage{tcolorbox}
\usepackage{xcolor}
\usepackage{soul} 
\usepackage{enumerate}
\usepackage{array}
\usepackage{multirow}
\usepackage{amssymb}
\usepackage{booktabs} % add in preamble
\usepackage{tabularx} % for auto-width tables
\usepackage{overpic} % for auto-width tables
\BibtexOrBiblatex

\newcommand{\deleted}[1]{}
\definecolor{lightblue}{RGB}{173, 216, 230}
\definecolor{lightgreen}{RGB}{197, 224, 180}
\definecolor{lightorange}{RGB}{255, 230, 153}
\definecolor{lightpink}{RGB}{255, 204, 204}
\definecolor{lightyellow}{RGB}{255, 255, 204}
\definecolor{lightpurple}{RGB}{230, 190, 255}
\definecolor{lightred}{RGB}{177, 80, 63}

\newtcolorbox{codefigure}[2][]{
    colback=white,
    colframe=darkgray,
    boxrule=0.5pt,
    arc=5pt,
    left=5pt,
    right=5pt,
    top=5pt,
    bottom=5pt,
    boxsep=0pt,
    title=#2,
    fonttitle=\large\bfseries,
    coltitle=white,
    colbacktitle=darkgray,
    before title={\vspace{5pt}},
    after title={\vspace{5pt}},
    #1
}

\newcommand{\name}{SAGE\xspace}
\newcommand{\clipA}{\ensuremath{\text{C}_\text{A}}\xspace}
\newcommand{\clipB}
{\ensuremath{\text{C}_\text{B}}\xspace}
\newcommand{\nofusers}{26\xspace}

\newcommand{\ssb}{\textsc{Skatepark-Biker}\xspace}
\newcommand{\hc}{\textsc{Handkerchief-Cruise}\xspace}

\newcommand{\ct}{\textsc{Cab-Train}\xspace}
\newcommand{\cp}{\textsc{Castle-Palace}\xspace}

\newcommand{\hb}{\textsc{Helicopter-Boat}\xspace}
\newcommand{\hd}{\textsc{Horse-Dog}\xspace}
\newcommand{\sss}{\textsc{Surfer-lSpeedboat}\xspace}
\newcommand{\cc}{\textsc{Candy-Cloud}\xspace}

\electronicVersion
\PrintedOrElectronic

\usepackage{hyperref}
\usepackage{cleveref}

\graphicspath{ {paper/figures/png_figures/} }
\title[Structure-Aware Generative Video Transitions]%
      {\name: Structure-Aware Generative Video Transitions \\ between Diverse Clips}

\author{
    \Large{\href{https://kan32501.github.io/sage.github.io/}{sage.github.io}}\\
    \large{
    Mia Kan\textsuperscript{1} \hspace{5mm}
    Yilin Liu\textsuperscript{1,2} \hspace{5mm}
    Niloy J. Mitra\textsuperscript{\thanks{Corresponding author.}1,3}} 
    \vspace{3mm}
    \\
    \large{\textsuperscript{1}University College London \hspace{3mm}
    \textsuperscript{2}Autodesk Research  \hspace{3mm}
    \textsuperscript{3}Adobe Research
    }
}

\begin{document}

\teaser{
 \centering
 \includegraphics[width=\linewidth]{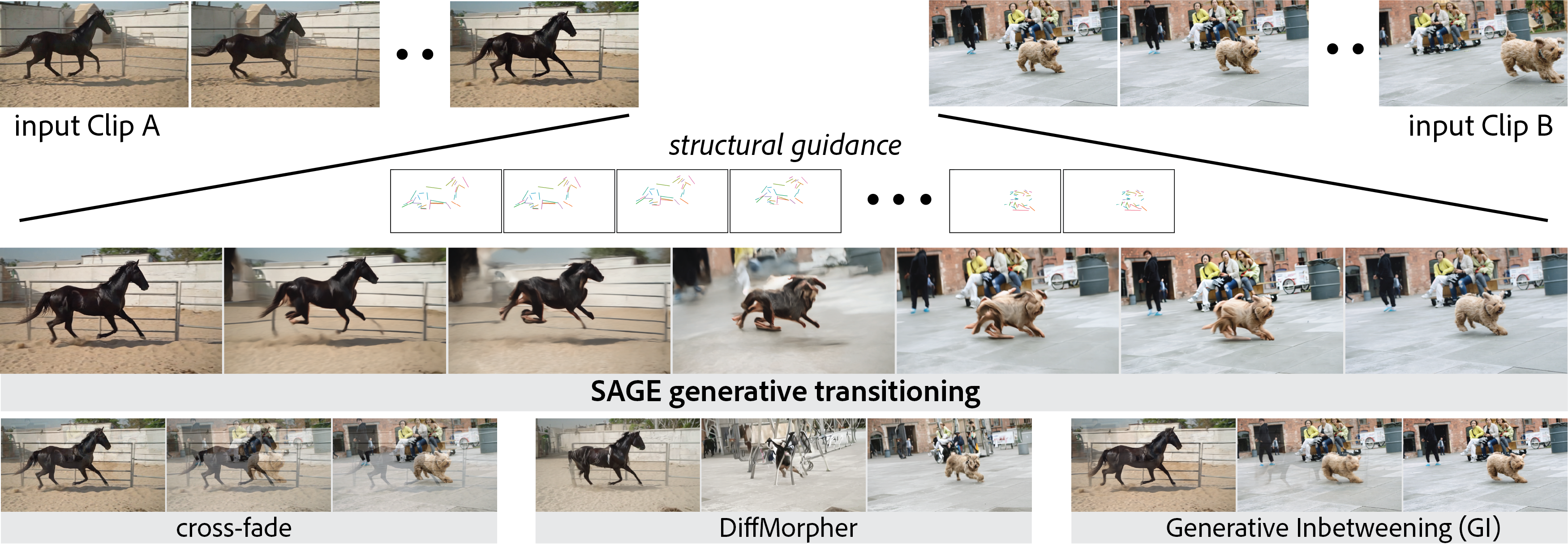}
  \caption{\textbf{\name: Structure-Aware Generative vidEo transitions.} 
Given two diverse clips~(top), \clipA and \clipB, prior approaches~(bottom) (e.g., cross-fade, DiffMorpher, Generative Inbetweening) often suffer from ghosting, structural collapse, or flicker. We introduce 
SAGE that extracts structural lines and motion cues, propagates them via B-spline trajectories to produce structural guidance. The guidance is then used to condition a pretrained diffusion model to synthesize temporally smooth with motion-coherent transitions~(middle) in a zero-shot setting.}
\label{fig:teaser}
}

\maketitle
\author{Mia Kan, Yilin Liu, Niloy J. Mitra}

%-------------------------------------------------------------------------

\begin{abstract}
Video transitions aim to synthesize intermediate frames between two clips, but naïve approaches such as linear blending introduce artifacts that limit professional use or break temporal coherence.
Traditional techniques (cross-fades, morphing, frame interpolation) and recent generative inbetweening methods can produce high-quality plausible intermediates, but they struggle with bridging \textit{diverse clips} involving large temporal gaps or significant semantic differences, leaving a gap for content-aware and visually coherent transitions.
We address this challenge by drawing on artistic workflows, distilling strategies such as aligning silhouettes and interpolating salient features to preserve structure and perceptual continuity.
Building on these strategies, we propose \name (Structure-Aware Generative vidEo transitions) as a simple yet effective zeroshot approach that combines structural guidance, provided via line maps and motion flow, with generative synthesis, enabling smooth, motion-consistent transitions without fine-tuning.
Extensive experiments and comparison with current alternatives, namely  \cite{film,tvg2024,zhang2023diffmorpher,vace,zhu2025generative}, demonstrate that \name outperforms both classical and the latest generative baselines on quantitative metrics and user studies for producing transitions between diverse clips.
The simple method effectively bypasses the need to acquire suitable training data, which is particularly difficult in our creative setting involving diverse clips. 
Code is available via the project page at \url{https://kan32501.github.io/sage.github.io/}.



\end{abstract}  
\section{Introduction}
\label{sec:intro}

%para 1: 
%Problem: why video transitions matter; where naive approaches fail
\textit{Video transition} refers to the task of synthesizing intermediate frames to seamlessly bridge two video clips. 
Such transitions are essential in editing, storytelling, and generative media, enabling fluid scene changes without distracting the viewer. 
Naïve strategies, such as linear blending in pixel or latent space, often introduce flickering, ghosting, or spurious objects, breaking temporal coherence and making them unsuitable for professional workflows.

%para 2:
%Gap: traditional methods vs. generative; what’s missing

More advanced methods, including morphing \cite{beier1992feature, wolberg1998image} and frame interpolation \cite{film, rife, dain}, treat transitions as interpolation between the start and end frames only. 
Recent generative inbetweening methods \cite{vidim, zhu2025generative, zhang2025eden}, designed specifically for video, demonstrate that modern video generation models, when trained with suitable data, can hallucinate plausible intermediates. 
However, these approaches assume small temporal gaps and closely aligned  semantics, and they often fail when clips differ significantly in content and/or style  -- video pairs we refer to as \textit{diverse clips}. 
Thus, current methods lack a way to generate transitions that are both \emph{content-aware} and \textit{visually coherent} across diverse scenarios, especially across significantly different clip pairs -- a scenario that is engaging because of its creative possibilities.

\begin{figure}[t!]
    \centering
    \includegraphics[width=\columnwidth]{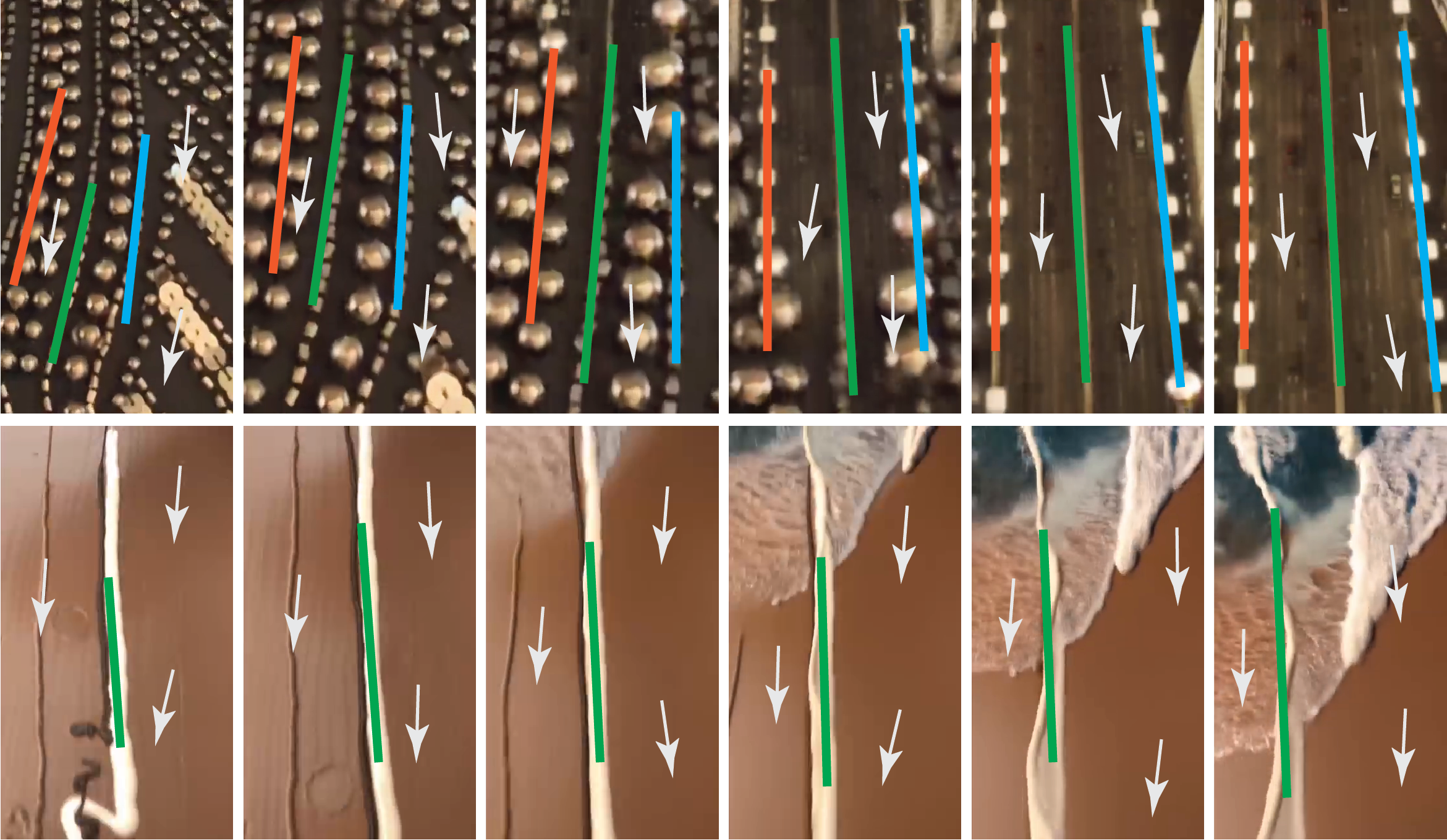} 
   \caption{\textbf{Artist-designed transitions.} \textit{
Two artist-crafted transitions illustrate the heuristics that inspire \name; full sequences are provided on the project page. 
(i) \emph{Structural anchoring}: silhouettes and edges are aligned across clips to prevent scene collapse, as highlighted by the matching colored lines.  
(ii) \emph{Motion continuity}: dominant flows such as camera pans are preserved to ensure fluid evolution, as indicated by the white arrows.  
(iii) \emph{Layered blending}: foreground objects morph while backgrounds fade, reducing ghosting and clutter (not depicted here).  
These principles motivate our design of structure- and motion-aware generative transitions.  }}
    \label{fig:motivation}
\end{figure}

%para 3:
%Our angle: artistic heuristics + why it’s challenging
We address this gap by targeting the challenging setting of diverse clips that differ in style, structure, and/or semantics. 
Skilled artists can craft compelling transitions in such cases, but rely on manual design tailored to specific pairs of clips (see \Cref{fig:motivation} and project page). Note that, as an additional challenge, we have very limited examples of such transitioning effects (e.g., in social media posts or creative content), and hence cannot finetune or retrain a generative model to directly produce such effects. 
Instead, we distill their heuristics into guiding strategies: first, detecting and aligning line features and silhouettes across clips, and second, interpolating salient features (e.g., feature lines and structural outlines) to anchor transitions in intermediate frames, as these provide smooth structural, semantic, and motion transitions. 
%As illustrated in \Cref{fig:motivation}, these heuristics motivate our algorithmic design. 

%para 4: 
%Solution: what we propose
We propose Structure-Aware Generative vidEo transitions (\name), a method that fuses \textit{structural guidance} with generative synthesis. 
Given a pair of video clips $\{\clipA, \clipB\}$, we first extract line maps and optical flow for the final frame of clip \clipA and the initial frame of clip \clipB. 
From these, we detect, match, and interpolate to produce intermediate line structures that capture both geometry and motion cues. Specifically, we demonstrate that suitably encoding detected linear structures, using their centers and slopes, along with the detected motion flows, allows us to establish better quality matches as well as produce more fluid motion-aware structural guidance. 
The extracted control structures can then be used to condition a pretrained generative inbetweening model~\cite{zhu2025generative}, producing temporally smooth and semantically consistent transitions, in a zero-shot fashion. 
Unlike prior methods, \name unifies geometric guidance with generative synthesis \textit{without} requiring fine-tuning, and makes use of both object appearance and motion while directly leveraging pretrained generative models.

%para 5:
%Contribution: what we show, how we evaluate
We assess \name across varied video transitions, benchmarking against classical interpolation (i.e., cross-fade) and state-of-the-art generative baselines (FILM \cite{film}, SEINE \cite{chen2023seine}, DiffMorpher \cite{zhang2023diffmorpher}, TVG~\cite{tvg2024}, Generative Inbetweening \cite{zhu2025generative}
and VACE \cite{vace}). 
Both quantitative results and a user study show that our method produces smoother, more natural transitions in the context of diverse clips. 
See \Cref{fig:teaser} and the project page for some examples.

\paragraph*{Novelty and contributions.} 
While our method is simple and leverages a pretrained generative inbetweening, the novelty lies in how these components are orchestrated to address a previously unexplored problem: \emph{content-aware video transitions across diverse clips in a zero-shot setting}. 
Unlike prior frame interpolation or generative inbetweening methods, which assume consistent semantics and small temporal gaps, we explicitly distill artist-inspired heuristics into a principled design. 
Specifically, we (i) introduce \emph{hierarchical structural anchoring}, where salient line structures are extracted, normalized, and matched in a layerwise manner to avoid background dominance; 
(ii) propose \emph{motion-aware B-spline propagation}, which couples local line evolution with global foreground trajectories, mitigating the trajectory crossings and incoherent motion seen in naive interpolation; and 
(iii) demonstrate that these structural priors can condition a pretrained diffusion-based inbetweening model to achieve smooth transitions without task-specific fine-tuning. 
This synthesis of structural guidance with generative synthesis is unique, enabling transitions that are both temporally coherent and semantically adaptive despite the absence of curated training data. 
%\textit{Code will be released upon acceptance. }

\section{Related Works}
\label{sec:related}

\paragraph*{Traditional video transitions.}
Video editing has long relied on handcrafted transitions such as cross-fades, wipes, and dissolves. 
These approaches, being procedural, are computationally simple and widely supported in editing software, but they are insensitive to scene content and therefore limited in realism and adaptability. 
Beyond preauthored transition functions, morphing techniques \cite{beier1992feature, wolberg1998image} represented an early attempt to interpolate structural correspondences between frames, but typically required manual keypoint alignment or feature specification, making them impractical for general-purpose video transitions. 
With the advent of generative models and access to suitable datasets, DiffMorpher \cite{zhang2023diffmorpher} set a new state-of-the-art by proposing diffusion-based generative morphing between images, providing a foundation for content-aware blending (see \Cref{sec:evaluation} for a comparison), though it does not leverage motion information present in video clips. 

\paragraph*{Video frame interpolation.}
Frame interpolation methods aim to generate intermediate frames between temporally adjacent inputs. 
Classical approaches estimated optical flow to warp pixels, while modern deep networks learn to predict motion or directly synthesize frames. 
Representative works include DAIN \cite{dain}, which leverages depth-aware warping; RIFE \cite{rife}, which predicts intermediate optical flows in real-time; and FILM \cite{film}, which targets large motion using multi-scale fusion with perceptual constraints (see recent survey \cite{acevfi}). 
While these methods achieve high-quality results for short-term interpolation, they typically assume that input frames share the same scene and semantics, and consequently fail when applied to transitions between diverse clips that differ substantially in appearance, motion, or style. 
Extensions such as SEINE \cite{chen2023seine}, which generates long videos with smooth and creative transitions between shot-level clips, and VACE \cite{vace}, which unifies video generation and editing tasks, broaden applicability but still operate primarily from keyframes and do not exploit motion cues for diverse clip-to-clip transitions. In \Cref{sec:evaluation} and the project page, we provide comparisons with FILM, SEINE, and VACE. 

\paragraph*{Generative inbetweening.}
Recent progress in diffusion-based generative models has enabled more powerful video synthesis. 
In particular, video diffusion models can effectively hallucinate plausible intermediate content beyond deterministic flow warping. For example, 
Jain et al.~\cite{vidim} propose a diffusion framework for video interpolation between frames; Zhou et al.~\cite{generativeInbetweening2024} adapt pretrained image-to-video diffusion models for keyframe interpolation using forward and backward temporal losses; Zhang et al.~\cite{zhang2025eden} enhance diffusion-based interpolation under large motion; and Zhang et al.~\cite{tvg2024} propose TVG as a training-free approach that interpolates in latent space using Gaussian process regression and frequency-aware fusion. 
These methods achieve strong performance when trained on large curated datasets, but generally assume small temporal gaps and semantic consistency. 
However, they are not directly applicable to artistic or cross-domain video transitions across diverse clips. In \Cref{sec:evaluation}, we present a comparison with a recent generative inbetweening approach \cite{generativeInbetweening2024} as well as TVG. 

\paragraph*{Concurrent work.}
Recent methods explicitly target transitions across clips with larger visual differences. MatchDiffusion \cite{pardo2024matchdiffusion} defines a \textit{joint} diffusion process to align coarse structure and motion for two different text prompts, followed by a \textit{disjoint} diffusion phase that generates detail-preserving, match-cut-ready video pairs. Although the results show broad structural coherence between the produced clips, they still require a manual blend step to bridge visual inconsistencies.
VTG~\cite{vtg2025} introduces a versatile diffusion-based framework\footnote{code was not publicly available at the time of publication.} for transitions, employing bidirectional motion fine-tuning and representation alignment, and evaluates on a dedicated benchmark (TransitBench). 
While promising, the method lacks fine structural control, which can lead to inconsistent motion or structural collapse in challenging cases. 
In contrast, our method integrates structural guidance (line structures and motion guidance) with pretrained generative inbetweening, enabling smoother and more semantically consistent transitions without additional training. 
\section{Design Considerations}
\label{sec:design_overview}

Artists and video creators often design video transitions manually, guided by heuristics that preserve perceptual continuity while allowing creative freedom. 
From examining such workflows from classical books~\cite{ondaatje1900conversations,pearlman2016cutting}, blog posts~\cite{descript2025videotransitions,vimeo2025masteringtransitions}, and popular social media examples, we distilled three principles that inform the design of \name. 

\paragraph*{(i) Structural anchoring.}
Transitions are smoother when dominant structural cues, such as edges, silhouettes, or perspective lines, are preserved across clips. 
Artists often align or morph these structures, even when the content changes significantly, to avoid abrupt scene collapses. 
This motivates our use of line maps and their motion encoding to provide structural guidance. 

\paragraph*{(ii) Motion continuity.}
Smooth transitions guide viewers' attention through consistent motion. 
Artists typically match or extrapolate dominant motion trajectories (e.g., camera pans, object flows, vanishing directions), ensuring that transitions feel fluid rather than chaotic and avoiding unnecessary crossings. 
This observation motivates our use of optical flow to preserve motion direction and magnitude, and blending using smooth B-spline paths to guide the intermediate transitions. 

\paragraph*{(iii) Layered blending.}
Manual transitions often separate global and local changes. For example, we observered that artists gradually faded backgrounds while foreground objects are interpolated or morphed. 
Such layering avoids ghosting and reduces visual clutter. 
Inspired by this, we design \name to combine structural and motion guidance with generative synthesis, while making use of foreground/background information; thus,  enabling both smooth global blending and coherent local transformations.

Following these principles, we design \name to extract structural lines and motion cues that anchor transitions across clips. 
By encoding these structures and using them to guide a pretrained generative inbetweening model, our approach operates in a zero-shot setting. 
This avoids the need for curated training data, which is scarce for artistic transitions across diverse clips.

\begin{figure*}
    \centering
\includegraphics[width=\linewidth]{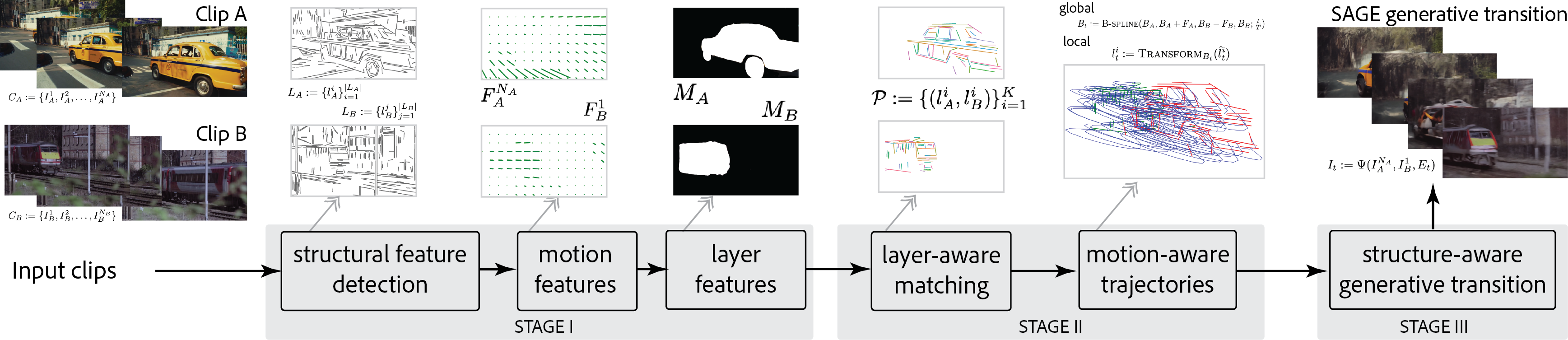}   \caption{
\textbf{Method overview.} \textit{Given two clips, we extract structural lines, optical flow, and foreground masks (Stage I). 
We match and interpolate these structures using motion-aware B-spline trajectories (Stage II), producing intermediate line sets $\{L_t\}_{t=1}^T$. 
These are then used to condition a pretrained generative inbetweening model (Stage III), yielding smooth and motion-aware transitions between diverse clips.}
}
    \label{fig:overview}
\end{figure*}

\section{Algorithm}
\label{sec:algorithm}

Given a pair of diverse video clips ${\clipA, \clipB}$, we extract $\{I_A^1, I_A^2, \ldots, I_A^{N_A}\}$ frames from $\clipA$ and $\{I_B^1, I_B^2, \ldots, I_B^{N_B}\}$ frames from $\clipB$. 
We aim to synthesize $T$ inbetween frames $\{I_t\}_{t=1}^{T}$ that are temporally smooth and semantically coherent across \emph{diverse clips}.
Guided by the design considerations in \Cref{sec:design_overview}, \name proceeds in three stages: 
(i) \textit{feature extraction} (\Cref{sec:feat}); 
(ii) \textit{motion-aware structural interpolation} (\Cref{sec:interp}); and 
(iii) \textit{conditional generative synthesis} (\Cref{sec:cond-gen}) ( see \Cref{fig:overview}). 
Our key contributions are the introduction of \emph{line-based structural extraction} to anchor transitions, and a novel \emph{B-spline–guided propagation} scheme that couples local line interpolation with global motion trajectories.

\subsection{Feature Extraction}
\label{sec:feat}
We extract three complementary features from the boundary frames $I_A^{N_A}$ and $I_B^1$:

\noindent (i) \textit{Structural features.} We detect two sets of line segments on the last frame $N_A$ of video clip \clipA and the first frame of video clip \clipB, where 
%\ylc{Do we want to mention we only do this on those particular two frames?}
\[
L_A := \{l_A^i\}_{i=1}^{|L_A|}, \quad L_B := \{l_B^j\}_{j=1}^{|L_B|},
\] 
using a pretrained line detector (we use GlueStick~\cite{PautratSYPL23}). 
Each line $l = \{x_1, y_1, x_2, y_2\}$ is encoded by its endpoints, collectively representing silhouettes and dominant contours.

\noindent (ii) \textit{Motion features.} We estimate optical flow fields $F_A^{N_A}$ and $F_B^1$ using SEA-RAFT~\cite{WangLD24}:
\[
F_A^{N_A} := \phi(I_A^{N_A-k}, I_A^{N_A}), 
\quad 
F_B^1 := \phi(I_B^1, I_B^k),
\]
with a small temporal span $k$, the number of offset frames used for optical flow computation. These capture local motion cues aligned with the structural features. Empirically, we set $k=3$ in our tests.
%\ylc{Do we still have to keep the notation $N_A$ and $N_B$ here, since we do not perform this operation on every frame but just on those particular boundary frames.}

\noindent (iii) \textit{Layer features.} Foreground masks $M_A, M_B$ are predicted with SAM~\cite{KirillovMRMRGXW23} with a user-specified click or coarse bounding box, isolating salient regions for line selection and preserving perceptual continuity.

%%%%%%%%%%%%%%%%%%%%%%%%%%%
\subsection{Interpolation via Structural Guidance}
\label{sec:interp}
To obtain a smooth yet content-aware transition between $\clipA$ and $\clipB$, we interpolate a sequence of intermediate structural primitives from the features in \Cref{sec:feat} by making use of the available motion cues. 
Specifically, we adopt a line-based interpolation scheme that produces intermediate line sets $\{L_t\}_{t=1}^T$ between $L_A$ and $L_B$, where $T$ is the number of inbetween frames. 
These line sets serve as geometric anchors that guide synthesis toward temporally coherent, semantically aligned transitions. 
A core design choice of \name is to propagate structural cues not by naïve linear blending, but by coupling (i) \emph{layer-aware line matching} and (ii) \emph{motion-aware B-spline trajectories}, thereby avoiding trajectory crossings and ensuring that interpolated structures respect both local geometry and global motion. 

%%%%%%%%%%%%%%%%%%%%%%%%%%%%%%%
\subsubsection{Layer-aware Line Matching}
We enforce structural consistency by performing \emph{layerwise} matching of lines using $L_A$, $L_B$ and the corresponding segmentation masks $M_A$, $M_B$. 
Directly matching all lines in $L_A$ to those in $L_B$ is brittle: background structures can dominate the objective and degrade the transition, particularly for \emph{diverse clips}. We therefore proceed in three steps, each reflecting a deliberate design decision:

\noindent
\emph{(i) Foreground selection.}  
We restrict attention to lines that lie in the foreground regions, selecting
\[
L_A^{\mathrm{fg}} := \{ l \in L_A \mid l \cap M_A \neq \emptyset \}, 
\quad 
L_B^{\mathrm{fg}} := \{ l \in L_B \mid l \cap M_B \neq \emptyset \}.
\]
This restricts subsequent marching on semantically salient objects while suppressing background clutter. 

\noindent
\emph{(ii) Canonical normalization.}  
For each foreground region, we compute a tight bounding box $B$ and normalize line endpoints into this canonical frame to obtain $\hat{L_A^{\mathrm{fg}}}$, $\hat{L_B^{\mathrm{fg}}}$. This makes matching robust to absolute position and scale, and ensures that correspondences are determined by relative geometric positioning rather than raw pixel coordinates.

\noindent
\emph{(iii) Hungarian matching.}  
We define a cost matrix $C \in \mathbb{R}^{|\hat{L_A^{\mathrm{fg}}}|\times|\hat{L_B^{\mathrm{fg}}}|}$ for all candidate line pairs, 
{
\[
C_{ij} := \|c(\hat{l}_A^i) - c(\hat{l}_B^j)\|_2^2 
\]
where $c(\cdot)$ is the canonical line center. Note that the cost matrix can be modified to include a weighted combination of other geometric attributes, such as line orientation and length. We prioritized predictable results for artists and opted for simpler position-only matching, as adding other features can cause line paths to contradict with each other later in the method.
% The weights $\alpha,\beta,\gamma$ balance spatial alignment, angular similarity, and scale. 
We use Hungarian matching to produce a set of one-to-one correspondences, 
\[
\mathcal{P} := \{(\hat{l}_A^i, \hat{l}_B^i)\}_{i=1}^K,
\] 
with unmatched lines discarded. These matched pairs $\hat{l}_A^i \leftrightarrow \hat{l}_B^i$ form the foundation for motion-guided interpolation, as described next.

\begin{figure*}[t]
    \centering
    \begin{overpic}[width=\linewidth]{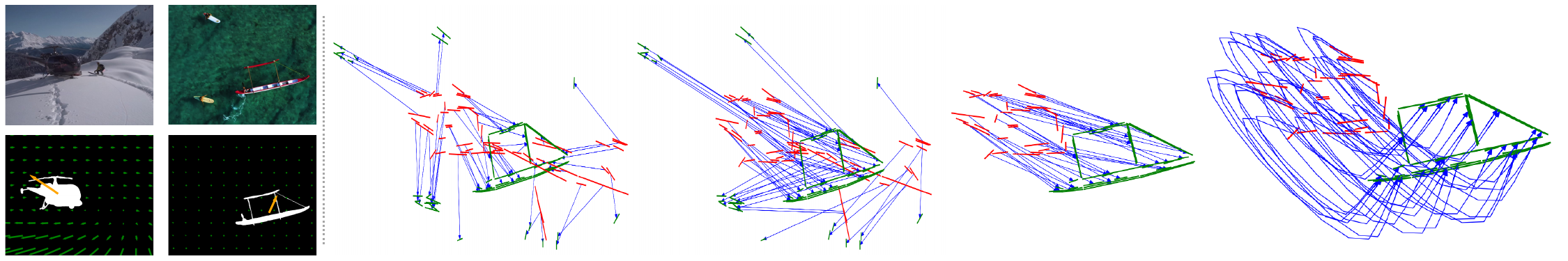} % , grid,tics=3
        \put(4.5,-1.5) {\footnotesize (a) Input frames}
        \put(23,-1.5) {\footnotesize (b) Full linear interpolation}
        \put(41.5,-1.5) {\footnotesize (c) w/ foreground/background}
        \put(62.5,-1.5) {\footnotesize (d) Only foreground}
        \put(79,-1.5) {\footnotesize (e) w/ B-spline interpolation}
    \end{overpic}
    \vspace{.01in}
    \caption{\textbf{Trajectory ablations for structural guidance.} \textit{
Trajectories (blue) shown from $L_1$ (red) to $L_T$ (green).
(a) Input frames with computed optical flow and segmentation; 
(b) Linear interpolation of \emph{all} detected lines, matched regardless of foreground/background, resulting in trajectory crossings and line mismatches due to neglect of semantic structure; 
(c) Linear interpolation of detected lines, matched by foreground/background, retaining undesired crossings and mismatches; (d) Linear interpolation restricted to \emph{foreground} lines (selected by $M_A$, $M_B$), yielding clearer trajectories for salient structures but still exhibiting crossover and motion inconsistency; 
(e) \emph{Motion-aware} guidance combining the global bounding-box trajectory $\{B_t\}_{t=1}^T$ with local line trajectories $\{L_t\}_{t=1}^T$, aligning structural evolution with scene/camera motion, as indicated in (a), while reducing trajectory crossovers. }
    }
    \label{fig:traj_ablation}
\end{figure*}

\subsubsection{Motion-aware B-spline Trajectories}
While one could interpolate each matched pair $(l_A^i, l_B^i) \in \mathcal{P}$ by blending direction, length, and center, such per-line interpolation ignores global scene dynamics. This often yields physically implausible trajectories (e.g., line crossings or structural collapse); see \Cref{fig:traj_ablation} and \Cref{sec:evaluation}. 
To address this, we introduce a two-scale interpolation scheme: first, a global foreground trajectory via B-splines; and then, local line blending within this evolving frame.

\emph{Global trajectory (B-spline guidance).}  
Foreground bounding boxes $B_A$ and $B_B$ are computed from $M_A$ and $M_B$. 
To capture dominant motion, we compute average flow vectors $F_A, F_B$ around matched lines and displace the bounding boxes accordingly. 
We then define control points $\{B_A, B_A+F_A, B_B-F_B, B_B\}$ and fit a cubic B-spline trajectory as, 
\[
B_t := \textsc{B-spline}(B_A, B_A+F_A, B_B-F_B, B_B; \tfrac{t}{T}), \quad t=1,\dots,T.
\]
This simple design ensures that interpolated structures follow a globally smooth and motion-aware path, effectively providing a local frame, avoiding abrupt jumps or unnatural crossings. 

\emph{Local line interpolation.}  
Given each pair, $\hat{l}_A^i, \hat{l}_B^i$, which are normalized in the canonical space of $B_A$ and $B_B$, intermediate lines are then blended linearly in the canonical space as, 
\[
\hat{l}_t^i := (1-\tfrac{t}{T}) \hat{l}_A^i + \tfrac{t}{T} \hat{l}_B^i,
\]
and mapped back into image space using the transformation defined by $B_t$:
\[
l_t^i := \textsc{Transform}_{B_t}(\hat{l}_t^i).
\]
We thus arrive at the resulting line sets,  
\[
L_t := \{l_t^i \mid (l_A^i, l_B^i) \in \mathcal{P}\}
\]
that encodes both local structure and global dynamics. 

\emph{Design rationale.}  
We chose this hierarchical strategy by combining B-spline foreground trajectories with local line interpolation to enforce smoothness of motion and semantic consistency simultaneously. Global trajectories capture camera/object motion, while local blending preserves fine structure. Together, they mitigate failures such as distracting line crossing, ghosting, or structural collapse that arise with naïve linear interpolation.

%%%%%%%%%%%%%%%%%%%%%%%%%%%
\subsection{Conditional Frame Generation} 
\label{sec:cond-gen} 
Finally, we condition a pretrained diffusion-based inbetweening model~\cite{zhu2025generative} with the interpolated line maps. 
The inbetweening model takes $(I_A^{N_A}, I_B^1, \{E_1 \dots E_T\})$ as input to produce intermediate frames $\{I_t\}_{t=1}^T$, 
\[
I_t := \Psi(I_A^{N_A}, I_B^1, E_t),
\]
where $\Psi$ denotes the generative diffusion sampler and $\{E_1 \dots E_T\}$ are edge maps that are passed as frame-wise conditions to the video generation model. 
The interpolated line maps are rasterized into frame-wise conditions by plotting each line set $\{L_t\}$ into an edge map $E_t$. 
These edge maps are then injected with $(I_A^{N_A}, I_B^1)$ via ControlNet-style conditioning. 
This enables zero-shot synthesis without fine-tuning, guided by structural and motion priors.

\section{Evaluation}
\label{sec:evaluation}

\begin{figure*}
    \centering
    \includegraphics[width=1\linewidth]{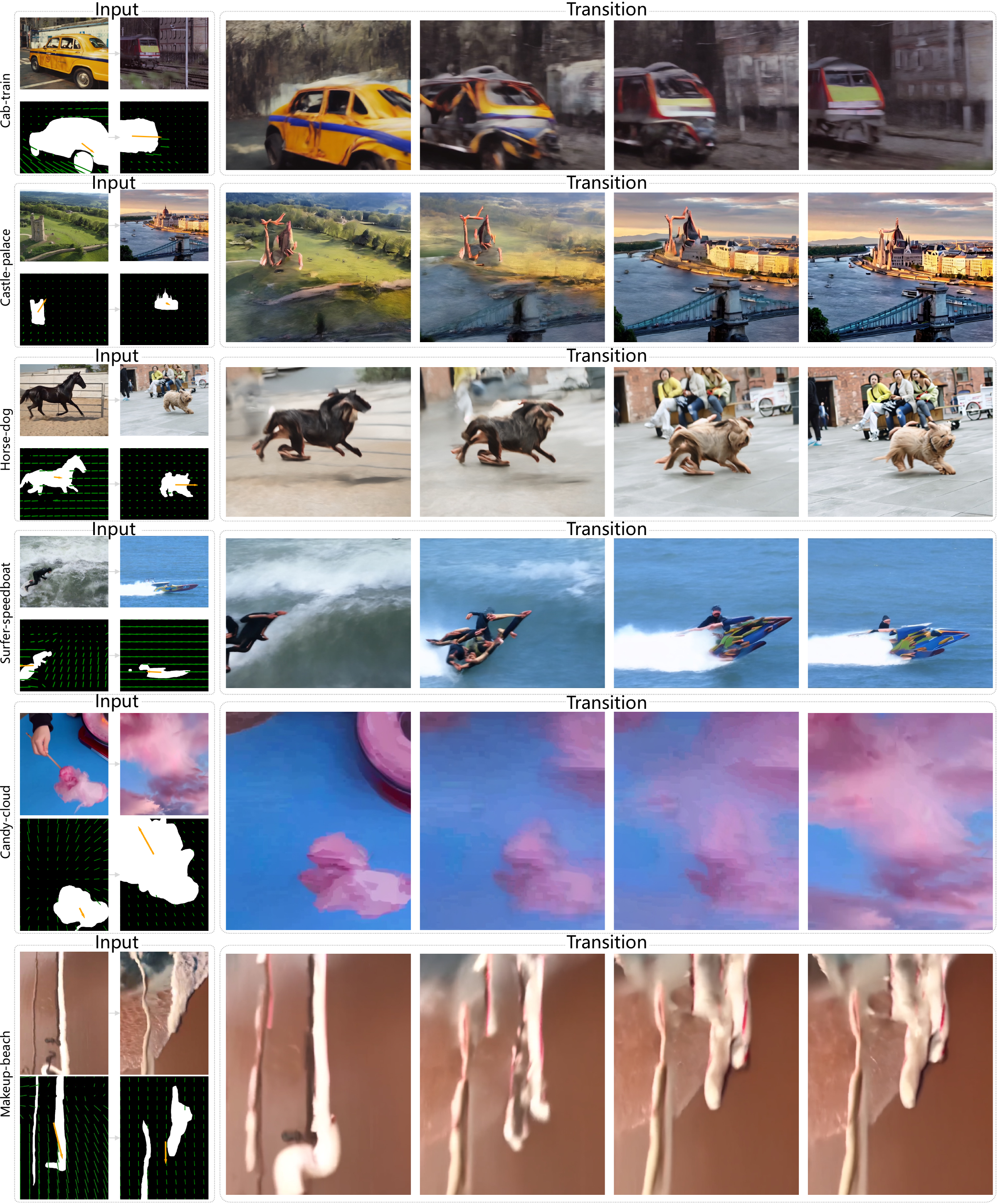}
    \caption{ 
    \textbf{Result gallery.} \textit{
        Qualitative results on \emph{diverse} video clips, showcasing the model's performance on complex transitions in scene scale (local-global), object category, and motion direction. Full videos are available on our project page.}
 }
    \label{fig:gallery}
\end{figure*}

\begin{figure*}
    \centering
    \includegraphics[width=0.98\linewidth]{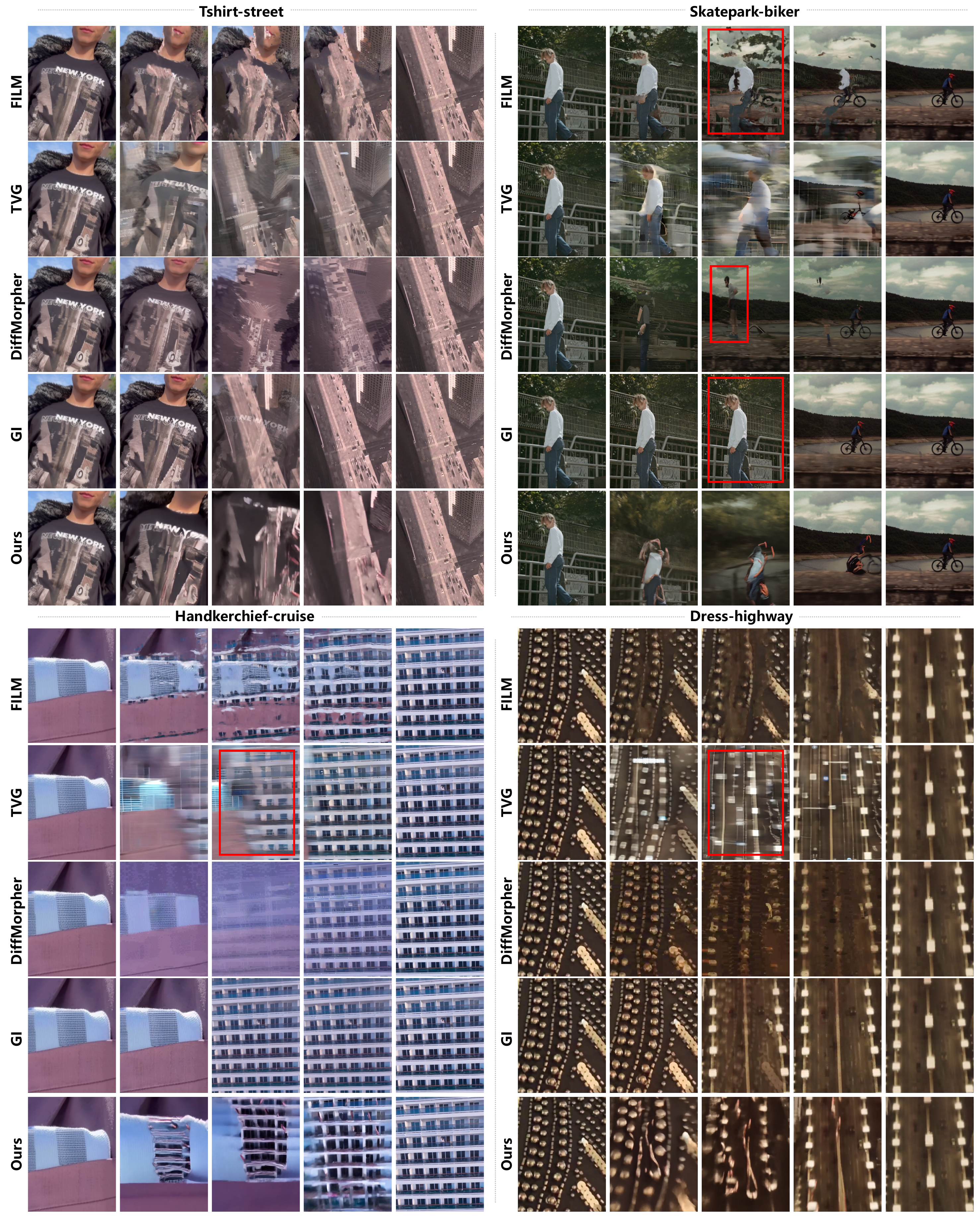}
    \caption{
    \textbf{Comparisons.} 
        \textit{Qualitative comparison with baseline methods, demonstrating that \name generates more plausible video transitions by maintaining consistency in motion, foreground objects, and background scenery. Full videos are available on our project page.}
    }
    \label{fig:comparison}
\end{figure*}

\paragraph*{Datasets.}
We evaluate \name and competing methods on a diverse set of video clip pairs. 
Our test set is drawn from three sources: (a)~artist-designed transitions, which provide high-quality reference examples; (b)~image pairs adapted from related work, where we generate short video clips using an image-to-video workflow; and (c)~diverse clips collected from public sources to span a wide range of motion, style, and complexity. 
The full dataset, along with reference transitions, will be released to enable reproducibility and facilitate future benchmarking. 

\paragraph*{Comparisons.}
We compare our approach against both classical and generative baselines. 
% As a procedural baseline, we include cross-fade. 
For content-aware morphing, we use DiffMorpher \cite{zhang2023diffmorpher}. 
Among generative transition methods, we test against FILM \cite{film}, TVG \cite{tvg2024}, and Generative Inbetweening \cite{generativeInbetweening2024} (GI). 
Finally, we also evaluate against VACE \cite{vace}, a universal video generation model capable of editing and synthesis (see project page). On the project page, we also qualitatively compare against SEINE~\cite{chen2023seine}. 

\paragraph*{Metrics.}
For the artist examples, we use the handcrafted transition as reference. 
We compute (i)~motion similarity, defined as the cosine similarity between optical flows extracted from the reference and generated transitions, and summed over the pixels; and (ii)~image and video similarity temporal smoothness metrics (FID, FVD). 
These allow us to capture both the motion smoothness and perceptual quality of transitions. 

\paragraph*{Implementation details.} We use SEA-RAFT~\cite{WangLD24} for computing optical flow, and normalize the results to unit vectors for similarity calculations. 
We take the transition videos generated by artists as ground truth videos when computing FID and FVD for the generated transition. 
% For \name, we use midpoint and slope encoding of line maps with a sampling resolution of 1 degree and 1 pixel for $\theta$ and $\rho$, respectively, and optical flow magnitude normalized to ... 
Unless stated otherwise, we use default hyperparameters across baselines. 
All method outputs are normalized to the same resolution and temporal length (13 in-between frames), for fair comparison.

\paragraph*{Runtime.}
The structural preprocessing is lightweight: on a single RTX 3090, feature and mask extraction take (1s) and line matching plus interpolation (2s); i.e., (\textasciitilde3s) total per transition. The runtime is then dominated by transition generation with the underlying diffusion model (5mins for 13 frames).

\paragraph*{Qualitative evaluation.}
Figure~\ref{fig:gallery} shows representative examples across domains, including artistic edits, object-centric scenes, and natural footage. 
Compared to interpolation methods, \name produces more coherent structure and consistent, smooth motion, 
even diverse motion direction (\ct and \sss), diverse scale changes (\cc), or diverse object category (\hd).
However, generative baselines occasionally collapse or introduce spurious content. 
For instance, FILM often results in a cross-fade-like transition, which can disrupt the fundamental structure and semantics of the figure (as illustrated in \ssb in \Cref{fig:comparison}).
Similarly, GI typically produces small, localized changes near the boundary frames, yet culminates in a sudden and abrupt mid-transition, indicating a failure to achieve a smooth temporal flow.
Although generative baselines are capable of producing transitions that are more semantically and structurally meaningful, their inherent lack of explicit structural prior yields undesirable artifacts. Examples of this include the introduction of an unrelated human figure in the \ssb result for DiffMorpher, and the consistent, distracting left-to-right wiping artifact observed in the outputs of TVG for both the \ssb and \hc examples, which disregard the actual motion direction. 
Additional examples are provided in the project page.

\begin{table}[b!]
    \centering
    \small
    \caption{
    \textbf{Quantitative comparisons.} 
    We report metrics on image quality (FID), video quality (FVD), and motion adherence (flow similarity). 
While some baselines achieve strong image/video scores (e.g., GI on FID, TVG on FVD), they fall short in preserving motion consistency. 
Our method attains the best flow similarity while remaining competitive in FID/FVD, demonstrating that a good transition must balance both visual fidelity and motion adherence rather than optimizing one at the expense of the other.
}
    \begin{tabularx}{.95\linewidth}{rccccc}
        \toprule
        Method & DiffMorp & GI & TVG & FILM & ours \\
        \midrule
        FID ↓ & \underline{151} & \textbf{147} & 157 & 157 & 153 \\
        FVD ↓ & 2641 & 2696 & \textbf{2093} & 2404 & \underline{2185} \\
        Flow similarity ↑    & \underline{0.61} & 0.55 & 0.57 & 0.56 & \textbf{0.69} \\
        \bottomrule
    \end{tabularx}
    \label{table:quantitative}
\end{table}

\paragraph*{Quantitative evaluation.}
\Cref{table:quantitative} presents a quantitative comparison against baseline methods. Our approach, \name, achieves the highest flow similarity to the ground truth (GT), which validates the use of motion consistency to constrain the transition. \name also secures the second-best results for FID and FVD. Note that, although GI obtains a better FID score, it simply duplicates boundary frames, creating an abrupt transition; as shown in \Cref{fig:comparison} and the project page. Similarly, TVG produces a constant left-to-right transition that fails to adapt to the source and target motion.

\paragraph*{User study.}
We conducted a user study to evaluate perceptual quality and preference. 
We recruited \nofusers participants with mixed backgrounds in video editing and graphics. 
Each participant was shown 24 pairs of transitions, comparing our method to one of the four baselines: DiffMorpher, Generative Inbetweening (GI), TVG, and FILM, randomly sampled from these sources. The order of our versus baseline methods was also randomized.  Each transition was concatenated with its corresponding input clips as ($C_A$, $T$, $C_B$) and was looped over unlimited time; the two concatenated clips were aligned side-by-side and played synchronously over time. \Cref{fig:user_study_screencap} shows a snapshot of the user study setup.

\begin{figure}[h!]
    \centering
    \includegraphics[width=\linewidth]{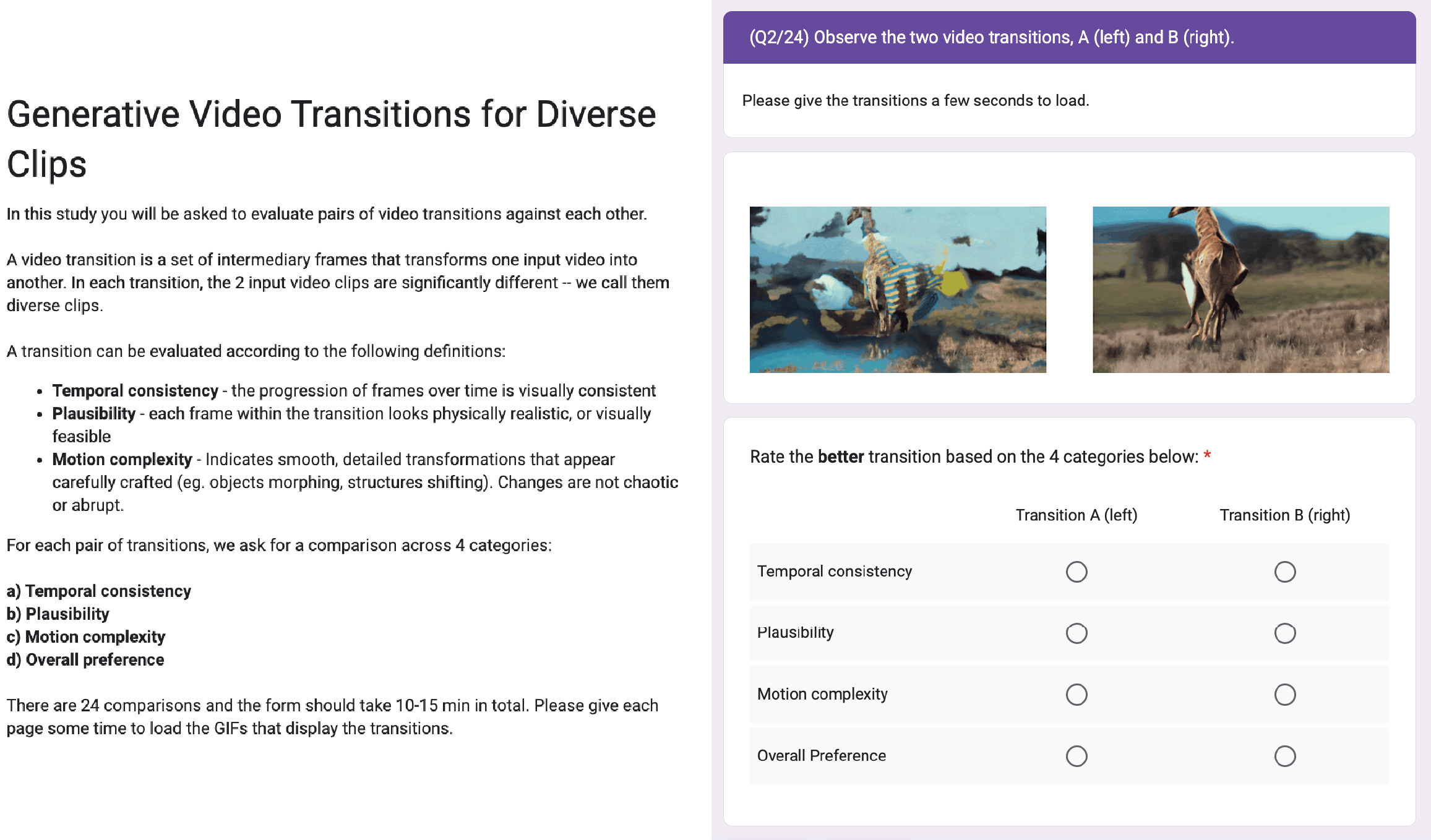}
    \caption{\textbf{User study design.} \textit{
    We conducted a user study to evaluate our method against multiple baselines. 
In each question, participants compared our approach with a randomly sampled baseline (FILM, TVG, DiffMorpher, or GI) across four criteria. 
(Left)~Starting instructions shown to participants; 
(right)~Example of a randomly sampled comparison pair.}
}
    \label{fig:user_study_screencap}
\end{figure}

\begin{table}[b!]
    \centering
    \scriptsize
    \caption{
    \textbf{User study results.}
    \name's video transitions were strongly preferred over all baseline methods. Users consistently rated our transitions higher across all measured aspects: \emph{Temporal Consistency}, \emph{Plausibility}, \emph{Motion Complexity}, and \emph{Overall Preference}. Note that a score of say x\% for method-y indicates that \name was preferred by x\% of the participants over method-y. 
    }
    \begin{tabularx}{.85\columnwidth}{rcccc}
        \toprule
        Method & FILM & TVG & DiffMorp & GI\\
        \midrule
        Temporal Consistency & 81.41\% & 80.12\% & 91.02\% & 86.53\% \\
        Plausibility         & 78.84\% & 76.28\% & 86.53\% & 80.12\% \\
        Motion Complexity    & 83.33\% & 82.69\% & 90.38\% & 89.10\% \\
        Overall Preference   & 82.69\% & 78.20\% & 89.10\% & 85.89\% \\
        Average              & 81.57\% & 79.33\% & 89.26\% & 85.42\% \\
        \bottomrule
    \end{tabularx}
    \label{tab:userstudy}
\end{table}

For each pair, participants were given a forced choice task between ours vs a baseline across four different criteria: (i)~which transition displayed better temporal consistency over time, (ii)~which transition displayed more plausibility, (iii)~which transition displayed more motion complexity (smooth, detailed changes), and (iv)~overall preference. 
Results (see Table~\ref{tab:userstudy}) show a clear preference for \name across all comparisons in each aspect.

% Qualitative feedback highlighted that \name avoided flickering and preserved motion direction more reliably than alternatives. 

\paragraph*{Ablations.}
We ablate the key components of \name and show results in the project page. 
Specifically, we evaluate: (i)~without structural guidance -- this is Generative Inbetweening~\cite{generativeInbetweening2024}; (ii)~without the layered  matching; and (iii)~using linear flow interpolation instead of B-spline interpolation for flow guidance. 
Results show that removing either structural or flow guidance significantly degrades transition quality, confirming the complementarity of both cues. Note that when the motions in the two source clips are already well-aligned, simple linear blending of matched structures performs comparably to our B-spline approach. However, for more diverse clip pairs where motion trajectories differ in direction, scale, or continuity -- B-spline interpolation produces smoother global paths and more natural motion inbetweening. Its effect is most visible in cases such as \ct, \ssb, and \hb, where clip motions are distinct; linear interpolation leads to sudden or inconsistent structural changes, whereas B-splines produce smoother, more anticipatory motion. It prevents abrupt movements and trajectory crossings that arise with linear blending. Together, these components account for the overall effectiveness of our method.  

\paragraph*{Failure cases.}
A limitation of our current approach stems from its reliance on a video generative backbone pretrained heavily on human poses. At times, this specificity can cause the model to interpret certain line configurations as limbs and hallucinate human-like structures along structural edges, as illustrated in the \cp and \sss examples of  \Cref{fig:gallery}. We believe that employing a more general-purpose video backbone would mitigate these artifacts – note that retraining on transition sequences is not an intuitive option due to the lack of suitable training data.

\paragraph*{Limitations.}
While \name demonstrates promising results for structure-aware generative transitions on diverse clips, it has several limitations. 
First, our approach relies on structural guidance from line maps and optical flow; when clips lack salient linear features or when flow estimation fails due to occlusions, textureless regions, or rapid motion, the resulting correspondences may be unreliable. 
Second, the method assumes that structural correspondences can be meaningfully established between clips; in highly abstract or stylistically divergent content, the extracted matches may be ambiguous or misleading. 
Finally, our current framework does not explicitly model appearance blending, which may lead to visual discontinuities in texture-rich regions. 
We believe these limitations can be addressed by extending our structural guidance with semantic cues (e.g., curved lines and/or Dino features), higher-order correspondence costs, and appearance-aware generation, which we leave for future work.

\section{Conclusion}
\label{sec:conclusion}

%summary
We have presented \name, the first method to produce structure-aware generative transitions between diverse video clips. 
Inspired by artist workflows, we demonstrated that carefully extracting, matching, and interpolating linear features across source clips provides effective structural guidance for generative video models. 
This design enables aesthetically pleasing and engaging transitions in a zero-shot setup, making it possible to realize generative workflows even in scenarios where collecting training data is infeasible.

%Future work
Looking forward, several promising directions remain. 
First, incorporating semantic cues (e.g., Dino features samples on the lines as in \cite{Dutt_2024_CVPR}) to score and refine feature matches could improve both robustness and perceptual quality. 
Second, integrating local smoothness priors and higher-order matching costs may further enhance correspondence estimation. 
Third, blending structural guidance with appearance information and motion flow could extend our framework into richer generative workflows (e.g., audio-guided~\cite{cheng2014imagespirit}), potentially combining structural interpolation with adaptive cross-fade strategies for even smoother transitions. 
We hope that our work provides both a practical tool for video editing and a foundation for future research on structure-aware generative transitions.

\paragraph*{Acknowledgments.}
We thank Burak Kizil for his assistance with the VACE comparisons, and Lilian Welschinger and Hao Xu for their valuable comments and suggestions. We are also grateful to the user study participants for their time, and to the anonymous reviewers for their constructive feedback on the original submission. This work was supported in part by generous gifts from Adobe Research and the UCL AI Centre.

%-------------------------------------------------------------------------
% bibtex
\bibliographystyle{eg-alpha-doi}  
\bibliography{generativeTransitioning}
% biblatex with biber
%\printbibliography                
% \onecolumn
% \onecolumn
% \newpage
% \twocolumn

% {\centering{\Large{\textbf{Supplementary Material for \name: Repurposing LLMs for Interior Designs}}}}

% \input{supplementary/supplemental}
% \bibliographystyle{eg-alpha-doi}  
% \bibliography{supplementary/roomGenerator_supp}

\end{document}